\documentclass{article}
\usepackage{float}

\usepackage{corl_2026}                 

\makeatletter
\providecommand{\@noticestring}{}
\makeatother
\usepackage{amsmath}
\usepackage{amssymb}
\usepackage{graphicx}
\usepackage{enumitem}
\usepackage{wrapfig}
\usepackage{svg}

\usepackage{microtype}                 
\usepackage{caption}
\usepackage{booktabs}
\usepackage{multirow}
\usepackage{colortbl}

\definecolor{deltagreen}{RGB}{30,130,30}   
\definecolor{marctint}{RGB}{232,244,234}    
\definecolor{basetint}{RGB}{242,242,242}    
\definecolor{headtint}{RGB}{236,239,244}    

\newcommand{\imp}[1]{\textcolor{deltagreen}{$\uparrow$\,#1}}   
\newcommand{\bestv}[1]{\textbf{#1}}                            

\usepackage{listings}
\usepackage{tcolorbox}
\tcbuselibrary{breakable, skins, listings}

\lstdefinestyle{inner}{
  basicstyle=\ttfamily,
  breaklines=true,
  breakatwhitespace=false,
  breakindent=0pt,
  frame=none,
  aboveskip=0pt, belowskip=0pt,
  columns=fullflexible,
}

\newtcolorbox{promptcard}[1]{
  title={\small\textbf{#1}},
  colback=gray!3, colframe=gray!25,
  coltitle=black,
  breakable,
  left=8pt, right=8pt, top=6pt, bottom=6pt,
  boxsep=0pt,
}

\title{Transferring Contact, Not Just Motion:\\ Compliant Grasping Across Dexterous Hands}

\author{
  Soofiyan Atar\thanks{Equal contribution.} \quad Yao-Ting Huang\footnotemark[1] \quad Michael Yip\\
  Department of Electrical and Computer Engineering\\
  University of California San Diego, United States\\
}

\begin{document}
\maketitle

\vspace{-1.0em}
{
\centering
\refstepcounter{figure}
\includegraphics[width=\linewidth, trim=0 148 0 148, clip]{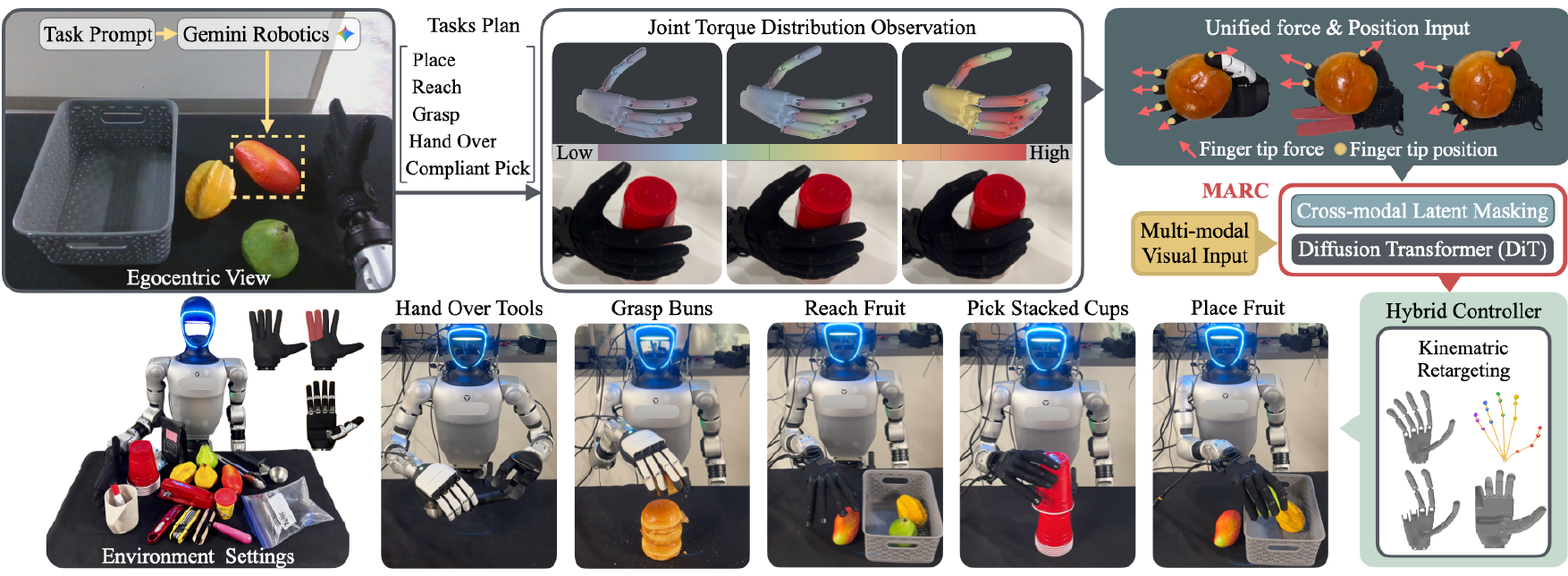}

{\small Figure~\thefigure:\ \textbf{Overview.} We propose a unified force--position encoding for dexterous hands that transfers across embodiments. A VLM grounds language instructions into task keypoints, and a state machine dispatches each phase to an optimization-based primitive, enabling stable long-horizon grasping without an end-to-end policy. The bottom row shows primitive actions across embodiments on compliant and rigid objects.}
\label{fig:cover_image}
\par
}

\vspace{-0.25cm}
\begin{abstract}
Dexterous grasping depends on contact regulation, not motion alone. Stable manipulation requires fingers to maintain appropriate object loading as contacts slip, deform, or become visually occluded. Existing cross-embodiment dexterous policies unify motion through retargeted hand poses or latent actions, but force feedback remains tied to each hand's sensing and actuation, limiting transfer. This work introduces a cross-embodiment force--position interface for contact-aware manipulation across heterogeneous dexterous hands. Motion intent is represented in a shared hand-pose latent, while each hand's effort signal is calibrated through system identification into physical joint torque in $\mathrm{N\!\cdot\!m}$. These torques are mapped to fingertip forces and compact per-finger load descriptors, giving the policy comparable observations of \emph{where} the hand should move and \emph{how} the object is loaded. Using this interface, a flow-matching visuomotor policy is trained on vision, proprioception, and calibrated contact, with structured visual masking that encourages reliance on force under grasp-relevant occlusion. The same calibrated signal drives a hybrid force--position controller for demonstration collection and execution, keeping force targets consistent across training and deployment. Experiments across structurally different hands show that calibrated contact feedback enables transferable compliant grasping, with learned primitives reusable in long-horizon manipulation pipelines.
\end{abstract}

\vspace{-1em}

\section{Introduction}
Grasping a fragile object, carrying it, and handing it to a person requires a robot hand not only to reach the right configuration, but also to regulate how strongly each fingertip loads the object as it slips, deforms, or becomes occluded by the hand itself. A policy that predicts only finger motion~\citep{yuan2024crossembodimentdexterousgraspingreinforcement, jiang2026crosshandlatentrepresentationvisionlanguageaction} can close on a plausible pose while remaining blind to whether the object is supported, over-constrained, or about to be dropped. Stable, compliant manipulation~\citep{Hogan_1985} therefore requires feedback on interaction force, not only on joint position.

Robotic hands frustrate this requirement because they differ in joint dimensionality, kinematic layout, actuation, and sensing, so behavior learned in one hand's native state--action space rarely transfers directly to another. Existing cross-embodiment methods reduce this gap by aligning motion through retargeting, contact maps, morphology embeddings, or latent action spaces, giving different hands a common command interface~\citep{yuan2024crossembodimentdexterousgraspingreinforcement,wu2025cedexcrossembodimentdexterousgrasp,bauer2026latentactiondiffusioncrossembodiment}. However, this unifies \emph{where} fingers move, not \emph{how} contact is loaded. The feedback that describes loading remains tied to hand-specific effort or tactile proxies, and vision alone provides weak recovery cues when contacts are self-occluded~\citep{bhirangi2024anyskinplugandplayskinsensing,zhang2025kinedexlearningtactileinformedvisuomotor}.

We make the contact itself transferable. Hand intent is represented through a shared pose latent, while each non-backdrivable or self-locking actuator's hand effort sensing is calibrated through system identification into physical joint torque, from which we derive fingertip forces and compact load descriptors. Unlike raw effort, torque and fingertip force express the interaction in physical coordinates that can be compared across hands. This yields a unified force--position interface in which both finger motion and grasp loading are expressed in a common set of transferable coordinates. Using this unified interface, the dexterous hand is driven by a hybrid force--position controller~\citep{seraji1987adaptive} used for teleoperation and autonomous rollout.

Built on this interface, we introduce MARC, a Mask-Aware Reactive Compliant flow-matching policy~\citep{mcallister2025flowmatchingpolicygradients} conditioned on vision, proprioception, and calibrated contact feedback. MARC uses structured visual masking during training so that the policy learns to rely on force and proprioception when grasp-relevant visual evidence is occluded. For long-horizon execution, MARC handles the contact-sensitive reach-and-lift phase, while model-based force controllers maintain the grasp during transport and release the object at handover. A high-level planner supplies task keypoints, but dexterous execution remains hand-agnostic and contact-driven.

The main contributions of this work are:
\begin{enumerate}[leftmargin=*,itemsep=2pt,topsep=2pt]
    \item a \textbf{transferable contact representation} that converts hand-specific effort and sensing proxies into calibrated joint torques, fingertip forces, and compact load descriptors, making contact feedback comparable across dexterous hands;
    \item a \textbf{unified force--position interface} that combines shared hand-pose intent with calibrated contact feedback, enabling transferable compliant control across heterogeneous dexterous hands;
    \item \textbf{MARC}, a mask-aware flow-matching policy trained on vision, proprioception, and calibrated contact, and a long-horizon execution pipeline that composes learned reach-and-lift behavior with model-based force controllers for transport and handover.
\end{enumerate}

Together, as shown in Fig.~\ref{fig:cover_image}, these components enable contact-rich manipulation skills to transfer across structurally different dexterous hands while maintaining stable, compliant interaction over long-horizon tasks.

\section{Related Work}
\textbf{Cross-embodiment dexterous manipulation} transfers skills across hands of different morphology using eigengrasp bases, contact maps, morphology embeddings, and latent action embeddings~\citep{yuan2024crossembodimentdexterousgraspingreinforcement,fei2025trograspefficientgraph,wu2025cedexcrossembodimentdexterousgrasp,zhang2026machagraspmorphologyawarecrossembodimentdexterous,bauer2026latentactiondiffusioncrossembodiment}. These reduce morphology gaps by aligning pose or action spaces, but do not make force feedback comparable in physical units. The closest learned-latent methods, XL-VLA and CrossDex~\citep{jiang2026crosshandlatentrepresentationvisionlanguageaction, yuan2024crossembodimentdexterousgraspingreinforcement}, fit a separate per-hand encoder/decoder from millions of teleoperated state--action pairs, repeated for every new hand. We instead use a single MANO-based~\citep{Romero_2017} hand-pose latent---where MANO is a parametric model of the human hand that encodes pose as a low-dimensional vector---learned once from roughly ten minutes of Manus glove~\citep{manus2025} motion rather than task demonstrations; retargeting maps it onto any hand, so no per-hand model is needed, resolving the kinematic pose mismatch and compressing the $63$-dimensional hand-pose input into a compact code. Our main departure is the contact channel: each hand's effort signals are calibrated into physical torque, yielding force features shared across embodiments.

\textbf{Scaling dexterous data.} A complementary line scales dexterous manipulation data through retargeting, human demonstration, and interaction capture, including DexMimicGen, DexWild, DexUMI, DEXOP, and DexFlyWheel~\citep{jiang2025dexmimicgenautomateddatageneration, tao2026dexwilddexteroushumaninteractions, xu2025dexumiusinghumanhand, fang2025dexopdevicerobotictransfer, zhu2025dexflywheelscalableselfimprovingdata, atar2026inhandmanipulationarticulatedtools}. Our work is orthogonal: we focus not on how demonstrations are collected, but on the representation in which they are expressed. Compliant dexterous control is well studied, yet many methods assume backdrivable or torque-controllable hands and are less applicable to non-backdrivable or self-locking actuators~\citep{shi2026minimalistcompliancecontrol}. Contact-rich dexterity has also used compliant grasp synthesis, tactile skins, and force- or touch-informed demonstrations~\citep{chen2024springgrasp, bhirangi2024anyskinplugandplayskinsensing, chen2026ptldsimtorealprivilegedtactile, zhang2025kinedexlearningtactileinformedvisuomotor, higuera2024sparshselfsupervisedtouchrepresentations, chen2026dexvitaccollectinghumanvisuotactilekinematic}. These methods usually encode contact as a sensor-specific modality. We instead use calibrated fingertip forces and torque-derived load descriptors as a unified contact representation, condition a Diffusion Transformer (DiT)~\citep{peebles2023scalablediffusionmodelstransformers} flow-matching policy on this force--position interface, and deploy it through a hybrid force/position controller with structured visual masking for self-occlusion.

\textbf{Robot foundation models} train generalist policies across heterogeneous robot datasets, language goals, and action spaces, including Open X-Embodiment, Octo, OpenVLA, $\pi_0$, RDT-1B, and DexVLA~\citep{embodimentcollaboration2025openxembodimentroboticlearning,octomodelteam2024octoopensourcegeneralistrobot,kim2024openvlaopensourcevisionlanguageactionmodel,black2026pi0visionlanguageactionflowmodel,liu2025rdt1bdiffusionfoundationmodel,wen2025dexvlavisionlanguagemodelplugin}. A related line uses VLMs for task planning, object grounding, and skill selection~\citep{liu2025robodexvlmvisuallanguagemodelenabled,zhong2025dexgraspvlavisionlanguageactionframeworkgeneral,debakker2026scaffoldingdexterousmanipulationvisionlanguage}. These improve semantic generalization but only align observations, language, and action formats, not a physically calibrated contact representation for transferring force feedback across hands. We therefore use VLMs only for high-level grounding and task decomposition, while execution is handled by a contact-aware policy in a shared hand-pose and calibrated force space.

\section{Preliminary}
\label{sec:preliminary}

\paragraph{Problem Formulation.}
\label{sec:problem_form}
We address compliant cross-embodiment dexterous manipulation from visual and contact observations. Let $\mathcal{H}$ be a set of hands, where each hand $h\in\mathcal{H}$ has $d_h$ actuated joints with native command $\mathbf{q}^{(h)}\in\mathbb{R}^{d_h}$, and differs in kinematics, actuation, and sensing. We remove this embodiment dependence through a unified command--contact interface. Commands are expressed in a shared MANO-pose latent $\boldsymbol{\psi}\in\mathbb{R}^{d_\psi}$ and mapped to native joints by a hand-specific retargeting map $\mathcal{R}_h:\boldsymbol{\psi}\mapsto\mathbf{q}^{(h)}$. Contact is expressed in physical units: a hand-specific predictor $\mathcal{G}_h$ maps raw effort proxies $\tilde{\boldsymbol{\tau}}^{(h)}$ and proprioceptive histories to joint torques $\boldsymbol{\tau}^{(h)}\in\mathbb{R}^{d_h}$ in $\mathrm{N\!\cdot\!m}$, from which the Jacobian gives fingertip forces $\mathbf{f}_t$ and spatial torque descriptors $\boldsymbol{\gamma}_t$.

The policy state is the unified tuple $\mathbf{s}_t=[\mathbf{x}_t,\mathbf{p}_t,\mathbf{f}_t,\boldsymbol{\gamma}_t,\boldsymbol{\psi}_t]$, where $\mathbf{x}_t\in SE(3)$ is the end-effector pose, $\mathbf{p}_t$ are fingertip positions, $\mathbf{f}_t$ the calibrated fingertip forces, and $\boldsymbol{\gamma}_t=[c,\rho,\ell]$ the spatial torque descriptors encoding where contact loads each finger. Given an $n$-step history of states and RGB-D observations with object-template crops, the policy predicts an $H$-step action chunk $\mathbf{a}_{t:t+H}\sim\pi_\theta(\cdot\mid\mathbf{O}_{t-n+1:t},\mathbf{s}_{t-n+1:t})$ in the shared command space, executing $n_a\le H$ steps before replanning. Hand identity selects only $\mathcal{G}_h$ and $\mathcal{R}_h$; it is never given to $\pi_\theta$.

\section{Method}

\subsection{Cross-Embodiment Interface}
\label{sec:interface}
The interface has two components, a shared command space and a calibrated contact channel, both supplied to the policy and the controller.

The command space is a MANO pose latent shared across hands. Following CrossDex~\citep{yuan2024crossembodimentdexterousgraspingreinforcement}, we encode the $63$-dim MANO~\citep{MANO:SIGGRAPHASIA:2017} hand-pose vector $\boldsymbol{\theta}_m\!\in\!\mathbb{R}^{63}$ (the stacked 3-D keypoints) with a nonlinear autoencoder, fit once to ten minutes of glove motion, into a latent $\boldsymbol{\psi}\!\in\!\mathbb{R}^{d_\psi}$ ($d_\psi=15$) in which the policy acts, replacing the original PCA basis. The dimensionality $d_\psi=15$ was chosen to balance reconstruction accuracy against compactness. The retargeting map $\mathcal{R}_h$ realizes $\boldsymbol{\psi}$ on hand $h$ by decoding $\hat{\boldsymbol{\theta}}_m=D(\boldsymbol{\psi})$ through the decoder $D$ and solving a modified DexPilot retargeting~\citep{handa2019dexpilotvisionbasedteleoperation, wuji2026retargeting} for the joint configuration $\mathbf{q}^{(h)}$,
\begin{equation}
\mathbf{q}^{(h)}=\arg\min_{\mathbf{q}}\;\sum_{i}\bigl[\alpha_i L^{\mathrm{tip}}_i+(1-\alpha_i)L^{\mathrm{full}}_i\bigr]+\lambda\|\mathbf{q}-\mathbf{q}_{\mathrm{prev}}\|^2 .
\label{eq:retarget}
\end{equation}
We modify DexPilot's key-vector objective into a blend of two Huber error terms. $L^{\mathrm{tip}}_i$ measures the gap at finger $i$'s tip in both position and pointing direction, capturing precise pinches. $L^{\mathrm{full}}_i$ measures the same gap at three keypoints along finger $i$ (PIP, DIP, TIP), preserving overall finger shape. A per-finger weight $\alpha_i\!\in\![0,0.7]$ rises under detected pinch and blends the two, while $\lambda\|\mathbf{q}-\mathbf{q}_{\mathrm{prev}}\|^2$ regularizes for temporal smoothness. The embodiment-specific map is confined to $\mathcal{R}_h$, so a new hand is added from only its URDF, and decoded fingertip positions $\mathbf{p}_t$ also enter the policy state.

\begin{wrapfigure}{r}{0.45\linewidth}
    \centering
    \includegraphics[width=0.65\linewidth, trim=60 90 100 90, clip]{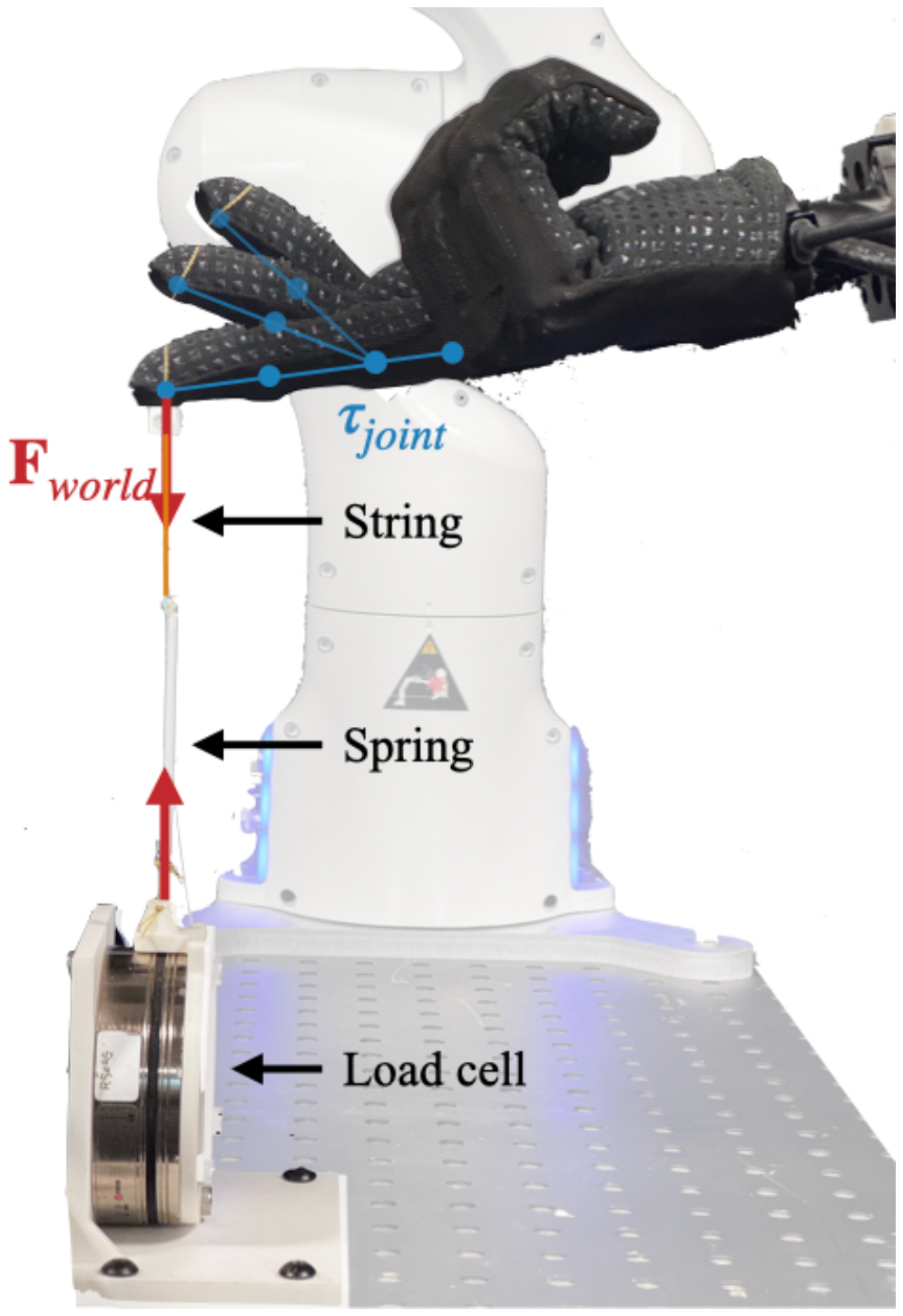}
    \caption{\textbf{Calibration setup.} Fingertip tethered through a string and spring to a six-axis load cell for per-hand torque calibration.}
    \label{fig:calibration}
\end{wrapfigure}

The contact channel calibrates the per-hand torque predictor $\mathcal{G}_h$ against a physical reference (Fig.~\ref{fig:calibration}). Each hand is mounted on a Franka at known poses with one fingertip tethered to an ATI Axia80 six-axis sensor by an inextensible string, so the transmitted load is a pure force $\mathbf{F}\!\in\!\mathbb{R}^{3}$ and the ground-truth torque of that finger's $n_j$ joints follows from its translational Jacobian $J_{\mathrm{lin}}\!\in\!\mathbb{R}^{3\times n_j}$ as $\boldsymbol{\tau}=J_{\mathrm{lin}}^{\!\top}\mathbf{F}$, with no moment term; these labels supervise $\mathcal{G}_h$. From the predicted torque $\boldsymbol{\tau}^{(h)}$ we read two per-finger quantities that stack over fingers into the state entries $\mathbf{f}_t$ and $\boldsymbol{\gamma}_t$. Writing $\boldsymbol{\tau}\!\in\!\mathbb{R}^{n_j}$ for a finger's entries of $\boldsymbol{\tau}^{(h)}$, its fingertip force is the least-squares inverse of $\boldsymbol{\tau}=J_{\mathrm{lin}}^{\!\top}\mathbf{F}$,
\begin{equation}
\mathbf{f}=\bigl(J_{\mathrm{lin}}J_{\mathrm{lin}}^{\!\top}\bigr)^{-1}J_{\mathrm{lin}}\,\boldsymbol{\tau},
\label{eq:fingertip_force}
\end{equation}
which is well defined whenever $J_{\mathrm{lin}}$ has full row rank, but returns a single net force at the tip and cannot localize where on the finger the contact occurred. The joint-torque vector preserves this localization, since each joint receives torque only from contacts distal to it along the chain. A contact on the proximal phalanx therefore loads only the base joint, a contact at the fingertip loads every joint, and the distribution of $|\tau_j|$ across the finger encodes the contact location. We summarize this distribution with three scalars normalized by the total absolute torque $s=\sum_j|\tau_j|+\varepsilon$,
\begin{equation}
c=\frac{\sum_j j\,|\tau_j|}{s},\qquad
\rho=\frac{s^2}{\sum_j \tau_j^2+\varepsilon},\qquad
\ell=\frac{|\tau_{\mathrm{lat}}|}{s},
\label{eq:contact_descriptors}
\end{equation}
the distal centroid $c$ (the torque-weighted mean joint index with $j$ ordered proximal to distal), the participation ratio $\rho\!\in\![1,n_j]$ (measuring how many joints share the load), and the lateral share $\ell$ (the fraction carried by the abduction-axis torque $\tau_{\mathrm{lat}}$ rather than flexion). Being model-free and location-agnostic, \eqref{eq:contact_descriptors} stays valid under non-fingertip contact, unlike \eqref{eq:fingertip_force}. Together, $\mathbf{f}_t$ and $\boldsymbol{\gamma}_t=[c,\rho,\ell]$ form the contact channel in physical, cross-hand-comparable units, giving the policy a compact spatial torque description of both the magnitude of fingertip loading and where along the finger that load concentrates.

\subsection{Compliant Dexterous-Hand Controller}
\label{sec:controllers}
The calibrated torque channel is not only a policy input but also the feedback for low-level execution. We wrap each hand command in a force-limited position controller that tracks the commanded pose under light load and retreats only the loaded joints once contact exceeds a target, applied identically during teleoperated demonstration and autonomous rollout.

The controller takes a position reference $\mathbf{q}_{\mathrm{ref}}\!\in\!\mathbb{R}^{d_h}$ from $\mathcal{R}_h$ or from the policy and corrects it with a one-sided retreat derived from the calibrated torque $\boldsymbol{\tau}^{(h)}$. An embodiment-invariant fingertip force target $\mathbf{f}_d$ maps to per-joint torque targets through the translational Jacobian as $\boldsymbol{\tau}_{\mathrm{des}}=J_{\mathrm{lin}}^{\!\top}\mathbf{f}_d$, giving consistent contact across kinematics. With gains $k_p,k_i,k_d$, retreat limit $o_{\max}$, and the convention that increasing $q_j$ closes joint $j$, the per-joint update is
\begin{align}
e_j&=|\tau_j|-\tau_{\mathrm{des},j}, &
I_j &\leftarrow \mathrm{clip}\!\big(I_j+e_j\Delta t,\,0,\,o_{\max}/k_i\big),\\[2pt]
r_j &= k_p\max(e_j,0)+k_i I_j+k_d\dot{e}_j, &
q_{\mathrm{cmd},j}&=q_{\mathrm{ref},j}-\mathrm{clip}(r_j,0,o_{\max}).
\end{align}
The correction is restricted to an opening motion because the proportional term acts only on overshoot, the integrator is floored at zero so that compliance relaxes when contact drops but never winds toward closure, and the output clip to $[0,o_{\max}]$ keeps every correction non-closing regardless of the derivative sign, while the $o_{\max}/k_i$ limit prevents integral windup. The command therefore stays within $o_{\max}$ of the reference, treats the reference as an upper bound on closure, leaves a lightly loaded grasp untouched, and settles where the measured torque equals its target.

Unlike a joint-space admittance law, whose bidirectional response can drive the hand deeper into the object, this controller can only release load. It leaves the policy commanding motion in the shared pose space while contact is regulated in calibrated torque, which is the action space of the flow-matching policy that follows.

\begin{figure}[t]
    \centering
    \includegraphics[width=0.8\linewidth, trim=1 35 1 35, clip]{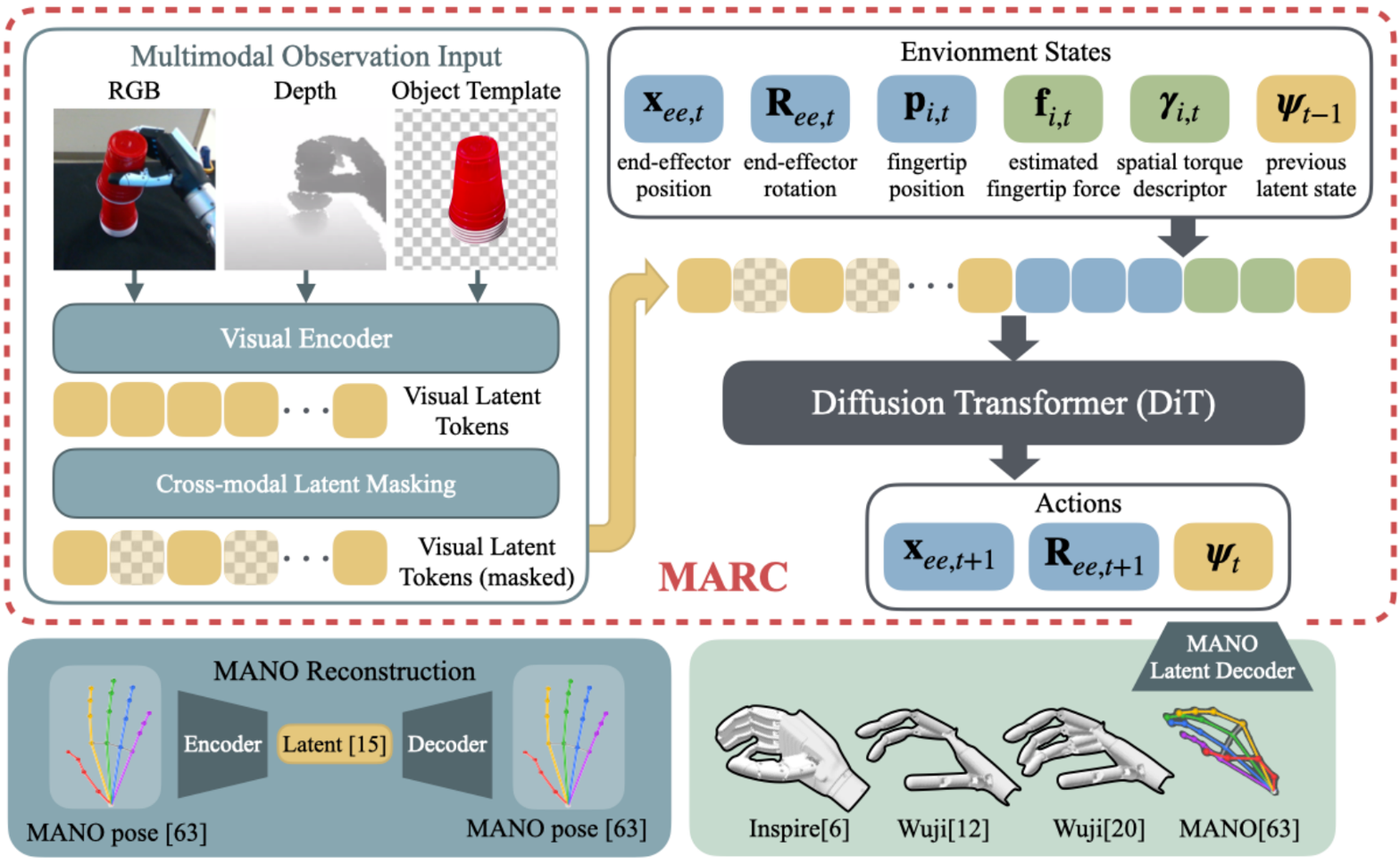}
    \caption{\textbf{Model pipeline.} MARC is a DiT flow-matching policy~\citep{mcallister2025flowmatchingpolicygradients} conditioned on a unified force--position state. The environment states (unified tuple $\mathbf{s}_t$) are as described in Sec.~\ref{sec:problem_form}. Structured masking is applied to visual patch tokens or entire camera streams during training, forcing the policy to act on force and proprioception under occlusion. MANO latent decoders are frozen, the DiT is trained from scratch, and the visual encoders are finetuned. At deployment, predicted actions are retargeted to each embodiment via its URDF.}
    \label{fig:marc}
\end{figure}

\subsection{MARC: Mask-Aware Flow-Matching Policy}
\label{sec:marc}
MARC realizes $\pi_\theta$ as a DiT-style flow-matching policy that predicts the $H$-step action chunk $\mathbf{a}_{t:t+H}$ from vision, proprioception, and calibrated contact observations as shown in Fig.~\ref{fig:marc}. Its key design is mask-aware conditioning: during training, visual patch tokens and entire camera streams are randomly replaced with learned mask embeddings, forcing the policy to rely on force and proprioceptive cues when visual evidence is missing or unreliable.

Each RGB, depth, and template stream is encoded by an ImageNet-initialized ResNet-18~\citep{he2015deepresiduallearningimage} with SpatialSoftmax pooling and projected into patch tokens. Masked visual tokens, together with the proprioceptive state $(\mathbf{x}_t,\mathbf{p}_t)$, contact features $(\mathbf{f}_t,\boldsymbol{\gamma}_t)$, and previous latent state $\boldsymbol{\psi}_{t-1}$, are concatenated into a conditioning vector $\mathbf{s}_t$. Given a demonstration action chunk $\mathbf{a}_1=\mathbf{a}_{t:t+H}$ (the target), Gaussian noise $\mathbf{a}_0$ (the source), and flow time $u\!\in\![0,1]$, MARC trains a velocity predictor $v_\theta$ on the linear flow
\begin{equation}
\mathbf{a}_u=(1-u)\mathbf{a}_0+u\,\mathbf{a}_1,\qquad
\mathbf{v}^{\star}=\mathbf{a}_1-\mathbf{a}_0,
\end{equation}
where $\mathbf{v}^{\star}$ is the ground-truth velocity transporting source to target, with objective
\begin{equation}
\mathcal{L}_{\mathrm{FM}}
=
\mathbb{E}_{u,\mathbf{a}_0,\mathbf{a}_1}
\bigl[
\|v_\theta(\mathbf{a}_u,u,\mathbf{s}_t)-\mathbf{v}^{\star}\|_2^2
\bigr].
\end{equation}
The action tokens are processed by DiT blocks whose normalization is modulated by flow-time and conditioning embeddings. At inference, an action chunk is generated by Euler integration of $v_\theta$ from Gaussian noise to $u=1$. When a camera is occluded or unavailable, its tokens are replaced with the same learned mask embeddings used during training, providing MARC with a unified sensor-availability interface rather than a separate failure-handling module.

\begin{wrapfigure}{r}{0.5\linewidth}
    \vspace{-1.5cm}
    \centering
    \includegraphics[width=1.0\linewidth, trim=0 100 0 100, clip]{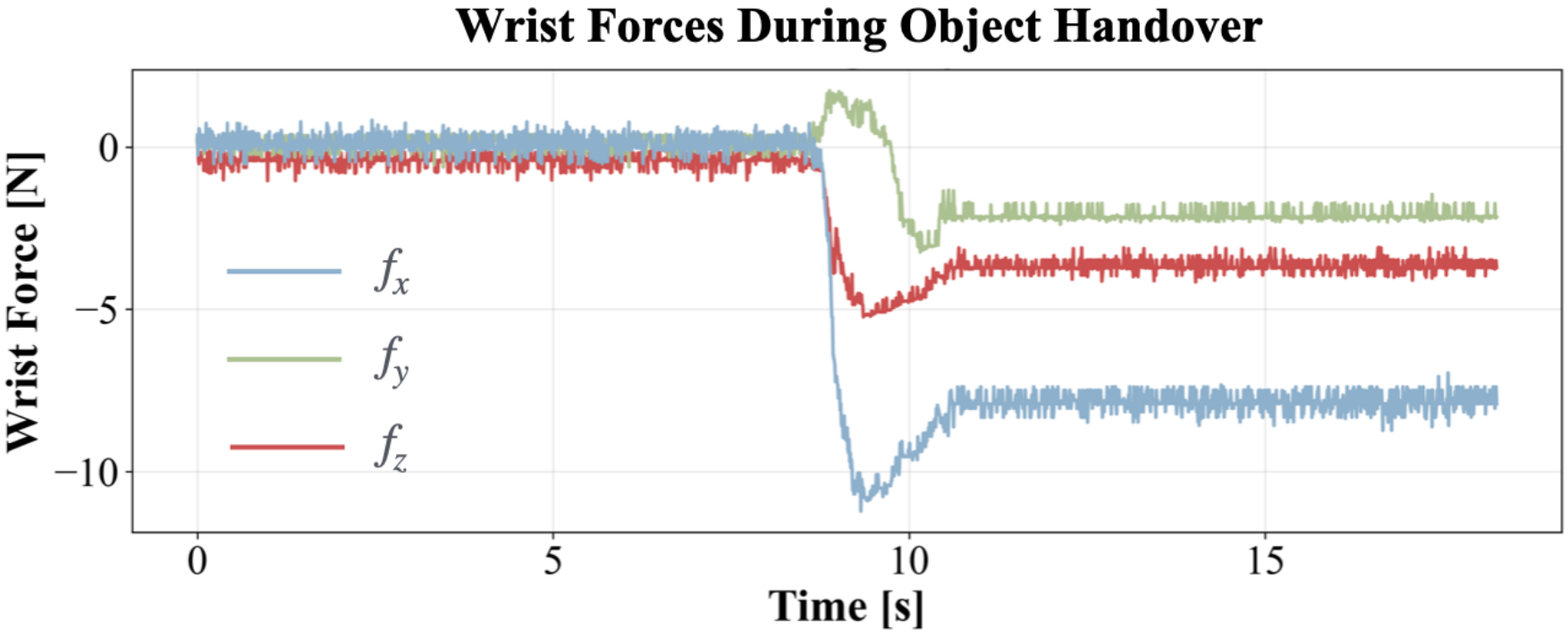}
    \caption{\textbf{Handover.} Wrist forces experienced during object handover.}
    \label{fig:handover}
    \vspace{0.6em}
    \centering
    \includegraphics[width=1.0\linewidth, trim=50 0 70 0, clip]{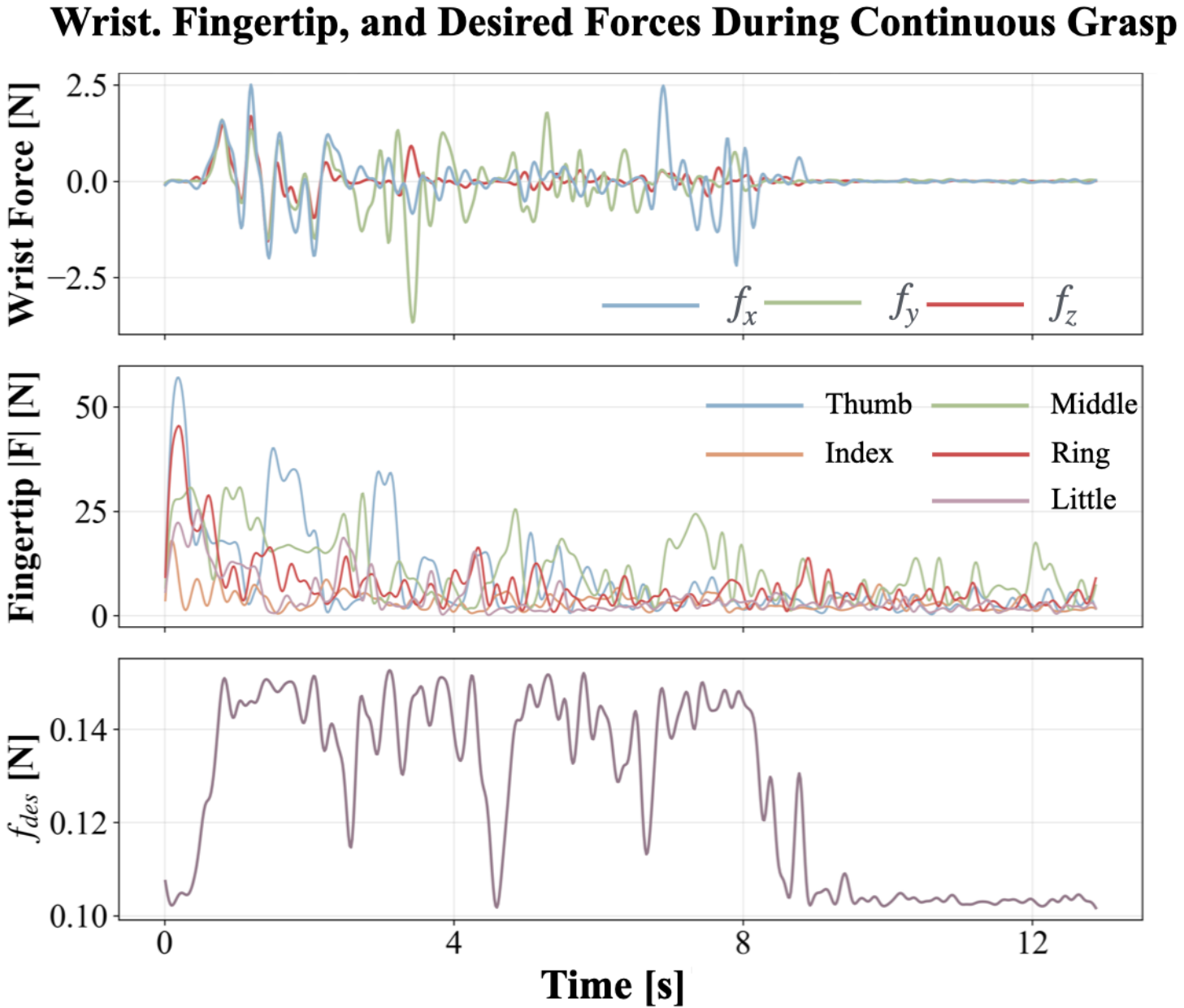}
    \caption{\textbf{Steady grasping.} Wrist force, fingertip force, and $\mathbf{f}_d$ during steady grasping for a compliant/rigid object.}
    \label{fig:continuous_grasp}
    \vspace{-2.5em}
\end{wrapfigure}

\subsection{Long-Horizon Task Composition}
\label{sec:longhorizon}
We compose long-horizon manipulation from reusable contact-aware primitives rather than a single end-to-end policy. A vision-language model proposes task keypoints from the scene image and provides only high-level spatial grounding instructions; it does not condition $\pi_\theta$. A deterministic state machine dispatches each phase to one of three primitives that share the command space of Sec.~\ref{sec:interface} but draw contact feedback from different sources.

\emph{Reach-and-lift} is executed by MARC, which performs the contact-sensitive grasp and initial lift toward the proposed keypoint. \emph{Transport} is handled by the steady-grasp regulator, which holds a minimal stable grip and increases closure as transport load registers in the calibrated hand torque $\boldsymbol{\tau}^{(h)}$. \emph{Handover} is handled by an end-effector admittance controller~\citep{ott2010unified} that yields the arm and opens the hand in proportion to the external wrench at the end effector. Transitions are gated by the calibrated contact, end-effector tolerance to the active keypoint, and handover progress. The same learned policy and model-based primitives thus cover lift, place, and handover without task-specific training.

\subsection{Optimization-Based Grasp and Handover Primitives}
\label{sec:primitives}
Two phases are handled by a continuous, model-based controller rather than the learned policy: holding the object during transport, and handover to a human or other arm in a bimanual setup. Both use only arm-joint proprioception and the calibrated contact signal shown in Fig.~\ref{fig:handover} and Fig.~\ref{fig:continuous_grasp}.

\textbf{Steady grasp.} During transport, the hand closes from the grasp pose $\mathbf{q}_g$ toward an adaptive grip target $\tau^{\star}+k_a\lVert\mathbf{a}\rVert$ that rises with arm acceleration $\lVert\mathbf{a}\rVert$, so the grip self-tightens exactly when transport loads the object:
\begin{equation}
\mathbf{q}_{\mathrm{cmd}}^{i+1}
= \mathbf{q}_g
+ \mathrm{clip}_{[0,\,\Delta q^{+}]}\!\Bigl(
  (\mathbf{q}_{\mathrm{cmd}}^{i}-\mathbf{q}_g)
  + \kappa\bigl(\tau^{\star}+k_a\lVert\mathbf{a}\rVert-\boldsymbol{\tau}_{i}\bigr)
\Bigr),
\label{eq:transport-grasp}
\end{equation}
where $\boldsymbol{\tau}_i$ is the measured finger torque at step $i$, $\kappa>0$ the gain, and $\tau^{\star}$ the resting grip. The clip to $[0,\Delta q^{+}]$ prevents loosening past the grasp and over-closure past the crush bound $\Delta q^{+}$. The arm carries the hand toward the goal via end-effector inverse kinematics.

\textbf{Handover.} A scalar progress state $\beta(t)\in[0,1]$ is integrated from the baseline-corrected wrist wrench $\mathbf{F}_\Delta\in\mathbb{R}^3$,
\begin{equation}
\dot{\beta}(t)=K\,\bigl[\,\|\mathbf{F}_\Delta(t)\|^2-F_{\mathrm{d}}^2\,\bigr]_+\,
\bigl(1-\beta(t)\bigr),\qquad\beta(0)=0,
\label{eq:beta-ode}
\end{equation}
where $K>0$ is the handover gain and $F_{\mathrm{d}}\geq 0$ a force deadband suppressing integration under noise or light incidental contact. The state $\beta$ simultaneously drives a finger release $\mathbf{q}_f=(1-\beta)\mathbf{q}_{g}+\beta\,\mathbf{q}_{\mathrm{open}}$ and gates a Cartesian admittance along the pull direction,
\begin{equation}
M_a\ddot{\mathbf{x}}_e+D_a\dot{\mathbf{x}}_e=\beta\,\mathbf{F}_\Delta,\qquad
\mathbf{x}_{\mathrm{cmd}}=\mathbf{x}_0+\mathbf{x}_e,
\label{eq:handover-admittance}
\end{equation}
where $\mathbf{x}_e$ is the end-effector offset from the nominal pose $\mathbf{x}_0$ and $(M_a,D_a)$ are the Cartesian inertia and damping. The release time-constant is thus set by the magnitude of the external wrench rather than by a fixed schedule. Full parameter listings are given in the Appendix.

\section{Experiments}
\label{sec:experiments}
\label{sec:setup}

\begin{wraptable}{r}{0.5\linewidth}
    \centering
    \vspace{-1.1cm}
    \caption{Dexterous hand comparison.}
    \label{tab:hardware}
    \small
    \renewcommand{\arraystretch}{1.15}
    \begin{tabular}{lccc}
        \toprule
        \textbf{Hand} & \textbf{\#Fingers} & \textbf{\#DoF} & \textbf{Mode} \\
        \midrule
        Wuji          & 5 & 20     & full actuation \\
        Wuji$_{12}$   & 3 & 12     & 2 fingers disabled \\
        Inspire       & 5 & 12 (6) & mimic coupling \\
        \bottomrule
    \end{tabular}
    \vspace{-0.25cm}
\end{wraptable}

\textbf{Tasks and data.} We evaluate on ten reach-and-lift tasks covering six rigid and four compliant or deformable objects. The rigid set comprises three toy fruits (mango, starfruit, guava), an ice-cream scooper, a pair of tongs, and a marker, while the compliant set comprises soft-dough cookies, brioche buns, stacked cups, and an egg. For each hand, we collect 50 teleoperated demonstrations, using Manus gloves for hand motion, a Vive Tracker for end-effector pose, and a foot pedal for clutching.

\textbf{Hardware.} All experiments run on a Unitree G1 humanoid fitted with three dexterous hands (Table~\ref{tab:hardware}); the same robot body and the same three embodiments are used across every task.

\textbf{Training and deployment.} Every model is trained on a single NVIDIA RTX 4090 for 60K steps at batch size 64. At deployment, policy inference runs on a laptop with an RTX 4070, with the VLM accessed via a cloud API and all low-level control and hand execution running locally on the same laptop. Unless otherwise stated, the training schedule, batch size, and inference pipeline are identical across all hand embodiments.

\subsection{Results}
\label{sec:results_reachlift}
We answer three questions about this pipeline: (1)~Does mask-aware conditioning on calibrated contact improve cross-embodiment reach-and-lift over a standard flow-matching policy? (2)~Is the calibrated force channel responsible for the gain? (3)~Does MARC generalize to an unseen hand configuration?

The evaluated objects span tools (ice-cream scoop, tongs), rigid items (toy fruits, marker), slippery and delicate compliant items (egg, brioche bun, soft-dough cookie), and a stability-critical compliant task (stacked cups), covering variation in grip type, friction, fragility, and required compliance. Two Wuji$_{12}$ cells warrant a note. The brioche-bun data collection was not done on Wuji$_{12}$ because its greasy surface demands a high-friction glove that the Wuji platform does not use, and Wuji$_{12}$'s two disabled fingers leave too little contact for a stable grasp. We kept the hand and gloves fixed across experiments to make every comparison apples-to-apples. The cookie scores zero for both policies on Wuji$_{12}$ because three fingers crumble its edges before stabilizing it, a precision regime that the 50-demonstration budget is too small to cover.

\textbf{Cross-hand comparison.} Table~\ref{tab:reachlift} shows flow matching policy with DiT backbone (sDiT) \cite{Yan2025ManiFlow, trilbmteam2025carefulexaminationlargebehavior} averaging $0.44$ across hands and tasks, dropping to $0.1$ on the Marker with Wuji. MARC lifts the overall mean to $0.71$ ($+0.27$) and improves every embodiment: Inspire $0.54\!\to\!0.84$, Wuji $0.48\!\to\!0.79$, Wuji$_{12}$ $0.29\!\to\!0.51$---with the smallest gain on the reduced embodiment whose two disabled fingers shrink the available contact area. The slim, contact-critical Marker improves the most (Inspire $0.2\!\to\!1.0$, Wuji $0.1\!\to\!0.9$), exactly the regime calibrated force feedback with visual masks was meant to address, given that occlusion relies more on force states. On deformables, the cross-hand mean rises from $0.43$ to $0.83$ for stacked cups and from $0.65$ to $0.90$ for brioche buns---objects whose acceptable grip window is narrow enough that a pose-only policy has no signal to regulate force below the crushing threshold. A subset comparison against Latent Action Diffusion~\citep{bauer2026latentactiondiffusioncrossembodiment} (Table~\ref{tab:latentdiff}) isolates the effect of MARC's contact-aware conditioning, since both methods share identical observation inputs and latent/action interfaces. MARC raises mean success from $0.19$ to $0.76$ ($+0.57$) and is strongest in every object--hand cell and per-hand mean, showing the gap is consistent across embodiments rather than driven by a single hand. The largest gain is on stacked cups ($0.00\!\to\!0.83$), where stability depends on force regulation rather than pose prediction alone.

\textbf{Effect of calibrated force.} Ablating the contact channel and retraining on rigid objects (Appendix) collapses the rigid mean from $0.75$ back to $0.48$, essentially the sDiT baseline. The gain, therefore, comes from the calibrated force channel and not from architecture or masking alone.

\begin{wrapfigure}{r}{0.5\linewidth}
    \vspace{-1.5em}
    \centering
    \includegraphics[width=\linewidth, trim=5 8 20 10, clip]{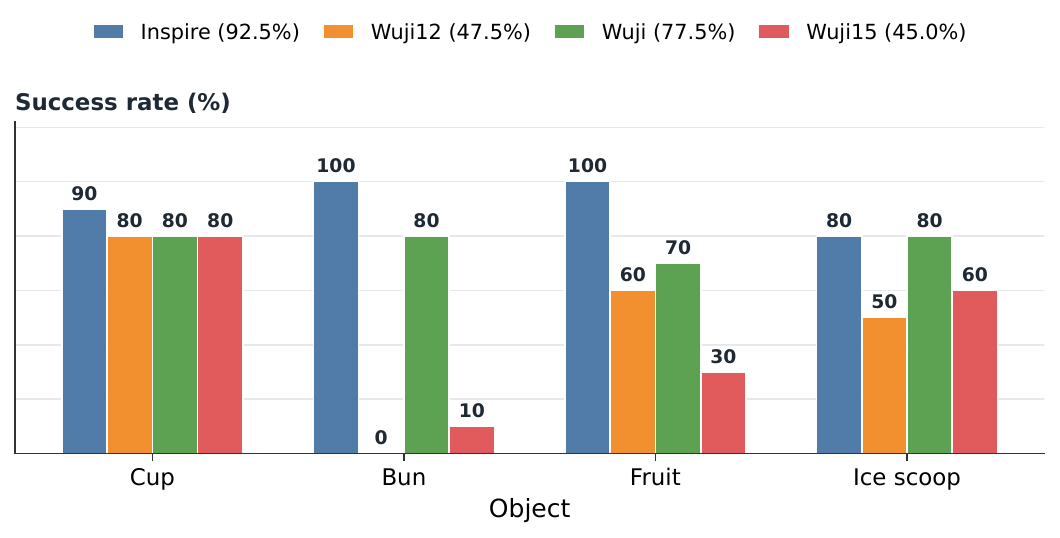}
    \caption{Success rates for seen and unseen robotic hands across objects. Bars show per-object success over 10 trials on MARC; legend values indicate mean success.}
    \label{fig:unseen}
    \vspace{-1.5em}
\end{wrapfigure}

\textbf{Unseen-hand generalization.} To test transfer beyond the training set (Fig.~\ref{fig:unseen}), we evaluate MARC on Wuji$_{15}$, a 15-DoF configuration not seen during training. Without retraining, success reaches $0.7$ on the ice-cream scoop, $0.6$ on the toy fruits, and $0.7$ on stacked cups, within $0.1$ of the in-distribution Wuji numbers, indicating that the shared MANO latent and the calibrated contact channel transfer across hand configurations rather than memorize a single configuration.

\begin{table*}[t]
\centering
\caption{\textbf{Reach-and-lift success rate} over 10 trials per object--hand pair: MARC vs.\ the DiT-flow baseline (sDiT). \textbf{Avg}: per-row mean over all evaluated objects. \emph{Mean}: cross-hand mean over hands with data. \textbf{$\Delta$ Mean}: MARC $-$ sDiT. ``---'' denotes not evaluated; \textbf{bold} marks the better method per cell.}
\label{tab:reachlift}
\small
\renewcommand{\arraystretch}{1.18}
\setlength{\tabcolsep}{5pt}
\resizebox{\textwidth}{!}{%
\begin{tabular}{ll cccc cccc c}
\toprule
 & & \multicolumn{4}{c}{\textbf{Rigid}} & \multicolumn{4}{c}{\textbf{Compliant / Deformable}} & \\
\cmidrule(lr){3-6}\cmidrule(lr){7-10}
\textbf{Method} & \textbf{Hand} & Fruit & Marker & Ice scoop & Tongs & Cups & Cookie & Bun & Egg & \textbf{Avg}\\
\midrule
\multirow{4}{*}{sDiT}
  & Inspire      & 0.9 & 0.2 & 0.6 & 0.4 & 0.8 & 0.2 & 0.6 & 0.6 & 0.54 \\
  & Wuji$_{12}$  & 0.3 & 0.3 & 0.4 & 0.5 & 0.1 & 0.0 & --- & 0.4 & 0.29 \\
  & Wuji         & 0.4 & 0.1 & 0.7 & 0.7 & 0.4 & 0.3 & 0.7 & 0.5 & 0.48 \\
\cmidrule(l){2-11}
  \rowcolor{basetint}
  & \emph{Mean}  & 0.53 & 0.20 & 0.57 & 0.53 & 0.43 & 0.17 & 0.65 & 0.50 & 0.44 \\
\midrule
\multirow{4}{*}{\textbf{MARC}}
  & Inspire      & \bestv{1.0} & \bestv{1.0} & \bestv{0.8} & \bestv{0.8} & \bestv{0.9} & \bestv{0.4} & \bestv{1.0} & \bestv{0.8} & \bestv{0.84}\\
  & Wuji$_{12}$  & \bestv{0.6} & 0.3          & \bestv{0.5} & \bestv{0.7} & \bestv{0.8} & 0.0          & ---          & \bestv{0.7} & \bestv{0.51}\\
  & Wuji         & \bestv{0.7} & \bestv{0.9} & \bestv{0.8} & \bestv{0.9} & \bestv{0.8} & \bestv{0.7} & \bestv{0.8} & \bestv{0.7} & \bestv{0.79}\\
\cmidrule(l){2-11}
  \rowcolor{marctint}
  & \emph{Mean}  & \bestv{0.77} & \bestv{0.73} & \bestv{0.70} & \bestv{0.80} & \bestv{0.83} & \bestv{0.37} & \bestv{0.90} & \bestv{0.73} & \bestv{0.71}\\
\midrule
\multicolumn{2}{l}{\textbf{$\Delta$ Mean}}
 & \imp{+0.24} & \imp{+0.53} & \imp{+0.13} & \imp{+0.27} & \imp{+0.40} & \imp{+0.20} & \imp{+0.25} & \imp{+0.23} & \imp{+0.27} \\
\bottomrule
\end{tabular}%
}
\end{table*}


\begin{table*}[t]
\centering
\caption{\textbf{Per-hand success rates: MARC vs.\ Latent Action Diffusion.}
Under matched observations and a shared latent/action interface, we report success over 10 rollouts per object--hand pair for three hands.}
\label{tab:latentdiff}
\small
\renewcommand{\arraystretch}{1.25}
\setlength{\tabcolsep}{8pt}
\vspace{-0.15cm}
\begin{tabular}{l ccc !{\color{black!25}\vrule} ccc}
\toprule
 & \multicolumn{3}{c!{\color{black!25}\vrule}}{\textbf{MARC}}
 & \multicolumn{3}{c}{\textbf{{Latent Action Diffusion}}} \\
\cmidrule(lr){2-4} \cmidrule(lr){5-7}
\textbf{Object} & Inspire & Wuji-12 & Wuji & Inspire & Wuji-12 & Wuji \\
\midrule
Fruit     & \bestv{1.00} & \bestv{0.60} & \bestv{0.70} & 0.30 & 0.30 & 0.30 \\
Marker    & \bestv{1.00} & \bestv{0.30} & \bestv{0.90} & 0.10 & 0.00 & 0.20 \\
Ice scoop & \bestv{0.80} & \bestv{0.50} & \bestv{0.80} & 0.40 & 0.10 & 0.20 \\
Tongs     & \bestv{0.80} & \bestv{0.70} & \bestv{0.90} & 0.30 & 0.30 & 0.20 \\
Cups      & \bestv{0.90} & \bestv{0.80} & \bestv{0.80} & 0.00 & 0.00 & 0.00 \\
Egg       & \bestv{0.80} & \bestv{0.70} & \bestv{0.70} & 0.40 & 0.20 & 0.10 \\
\midrule
\rowcolor{marctint}
\textbf{Mean} & \bestv{0.88} & \bestv{0.60} & \bestv{0.80} & 0.25 & 0.15 & 0.17 \\
\bottomrule
\vspace{-1cm}
\end{tabular}
\end{table*}

\section{Limitations and Future Work}
Our framework's transferability is bounded by the fidelity of the calibrated contact interface: every new hand requires per-finger system identification to recover joint torque from native effort, and residual calibration error propagates into both the controller's torque target and MARC's contact conditioning. Evaluation is confined to rigid and deformable tabletop objects; articulated objects, whose contact geometry and load distribution evolve under joint motion, remain untested. Finally, compliance gains $(k_p, k_i, k_d, o_{\max})$ are still hand- and task-specific, leaving automatic adaptation across objects and embodiments as future work.

\section{Conclusion}
\label{sec:conclusion}
We presented a cross-embodiment force--position interface that calibrates per-hand effort into joint torques, fingertip forces, and compact load descriptors, paired with a shared MANO-pose latent so that motion intent and grasp loading share transferable coordinates. Built on this interface, MARC is a mask-aware flow-matching policy conditioned on calibrated contact, executed through a hybrid force--position controller and composed with model-based steady-grasp and handover primitives. Across three structurally different hands and ten objects, MARC raises mean reach-and-lift success from $0.44$ to $0.71$ and transfers zero-shot to an unseen 15-DoF configuration, supporting calibrated contact as a practical axis of cross-embodiment transfer.


\bibliography{ref}

\clearpage
\clearpage
\section*{Appendix}
\setcounter{section}{0}
\setcounter{figure}{0}
\setcounter{table}{0}
\setcounter{equation}{0}

\renewcommand{\thesection}{S\arabic{section}}
\renewcommand{\thefigure}{S\arabic{figure}}
\renewcommand{\thetable}{S\arabic{table}}
\renewcommand{\theequation}{S\arabic{equation}}


\section{System Identification Details}
\label{app:sysid}

\paragraph{Mechanical setup.}
A Franka Emika Panda arm holds the dexterous hand at a fixed, precomputed pose. At the same time, an inextensible string connects the active finger's distal link to an ATI Axia80 six-axis F/T sensor mounted on a rigid load cell. Because the string transmits only a tensile force, the wrench at the sensor reduces to $\mathbf{F}_t \in \mathbb{R}^3$; moment components are discarded. All remaining fingers are held stationary throughout each trial.

\paragraph{Excitation waveforms.}
The active joint is driven through a randomly concatenated sequence of $10\,\mathrm{s}$ primitives, each at a uniformly sampled intensity and clamped to $[q_{\min}, q_{\max}]$:
\begin{equation}
  s(t) \;\in\;
  \Bigl\{
    \underbrace{at + b}_{\text{linear}},\;
    \underbrace{A\sin(2\pi f t + \varphi)}_{\text{sinusoid}},\;
    \underbrace{\mathrm{RW}(\sigma)}_{\text{random walk}},\;
    \underbrace{\mathrm{saw}(f)}_{\text{sawtooth}},\;
    \underbrace{\mathrm{stutter}(\Delta,p)}_{\text{stutter}},\;
    \underbrace{Ae^{\alpha t}}_{\text{exp.\ sweep}}
  \Bigr\}.
  \label{eq:excitation}
\end{equation}

\paragraph{Angular grid and logging.}
Trials are collected at equilibrium angles spaced $10^{\circ}$ apart across each joint's full range, with ten independent trials per grid point ($\approx\!100$ trials per $90^{\circ}$-range joint). The thumb uses a two-dimensional grid over its flexion and adduction--abduction axes jointly. The control loop runs at $20\,\mathrm{Hz}$, synchronously recording
\begin{equation}
  \mathcal{D}_t
  = \Bigl(
      q_t,\;
      \dot{q}_t,\;
      \tilde{\boldsymbol{\tau}}^{(h)}_t,\;
      \mathbf{F}_t^{(\mathrm{ATI})},\;
      \boldsymbol{\tau}^{(h)}_t
    \Bigr),
  \label{eq:log}
\end{equation}
where $\tilde{\boldsymbol{\tau}}^{(h)}_t$ are the native uncalibrated effort proxies and $\boldsymbol{\tau}^{(h)}_t = \mathbf{J}_{\mathrm{lin}}^{\top} \mathbf{F}_t^{(\mathrm{ATI})}$ the ground-truth torque labels. No-load trials (string absent) are merged into the training set to anchor the regressor's zero baseline.


\begin{figure}
    \centering
    \includegraphics[width=0.48\linewidth, trim=5 80 0 50, clip]{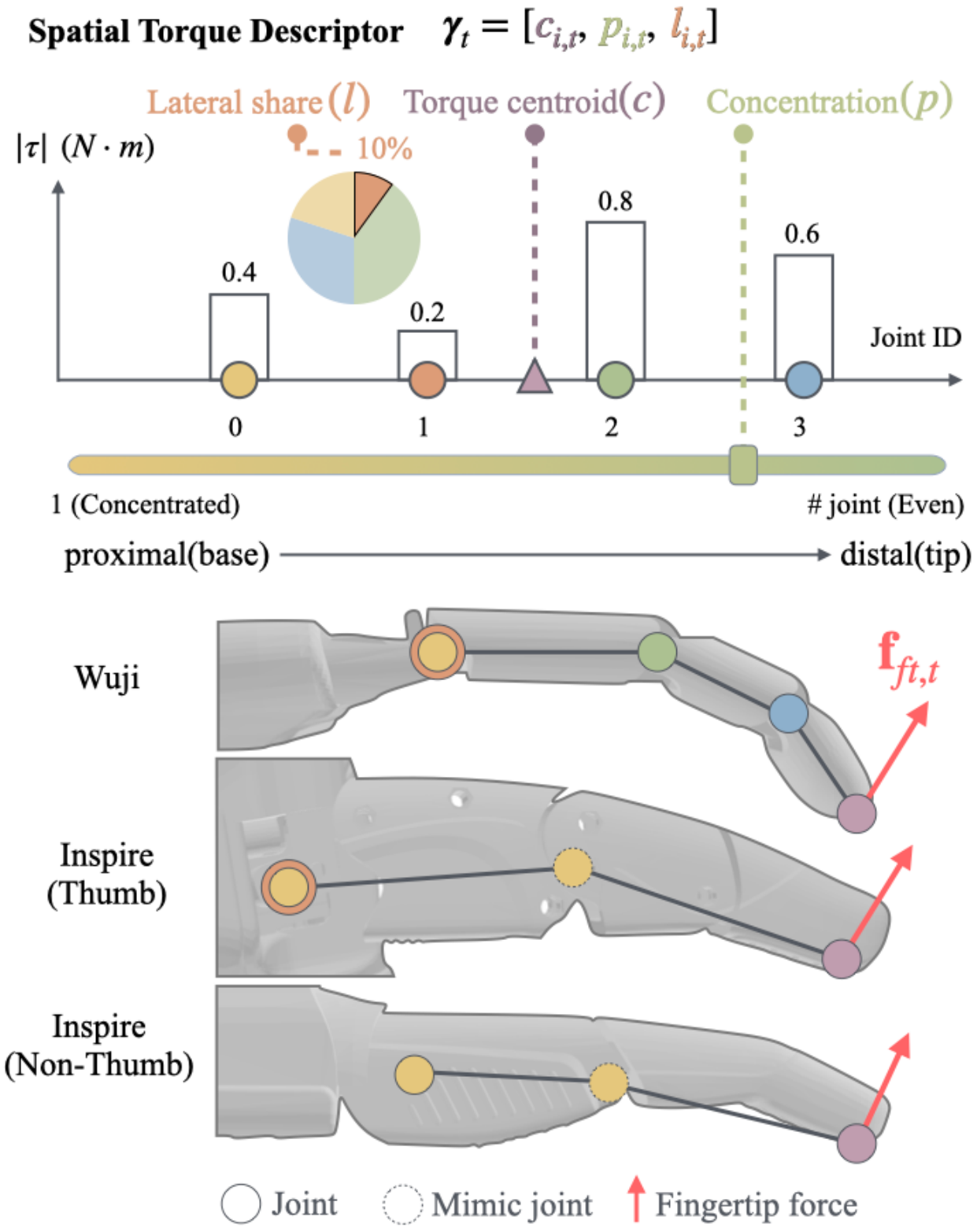}
    \caption{Spatial torque descriptor}
    \label{fig:spatial_torque_pred}
\end{figure}

\section{Torque Regressor: Architecture Selection}
\label{app:regressor}

\paragraph{Finger grouping and data pooling.}
Rather than training one regressor per finger, we exploit kinematic symmetry to pool data and share models within each motor group. For \emph{Inspire}, the four non-thumb fingers are identical, so their data are merged, and one model predicts a single torque ($d_{\mathrm{out}}{=}1$); the two-DoF thumb is a separate group whose model predicts both joint torques simultaneously ($d_{\mathrm{out}}{=}2$). For \emph{Wuji}, joints j3/j4 use the same actuator across all fingers (including thumb), so their data are pooled into one model ($d_{\mathrm{out}}{=}1$ per query joint); joints j1/j2 form a second group whose model ingests the concatenated feature sequences of both joints and outputs two torques ($d_{\mathrm{out}}{=}2$).

\begin{figure}
    \centering
    \includegraphics[width=0.48\linewidth, trim=50 50 0 50]{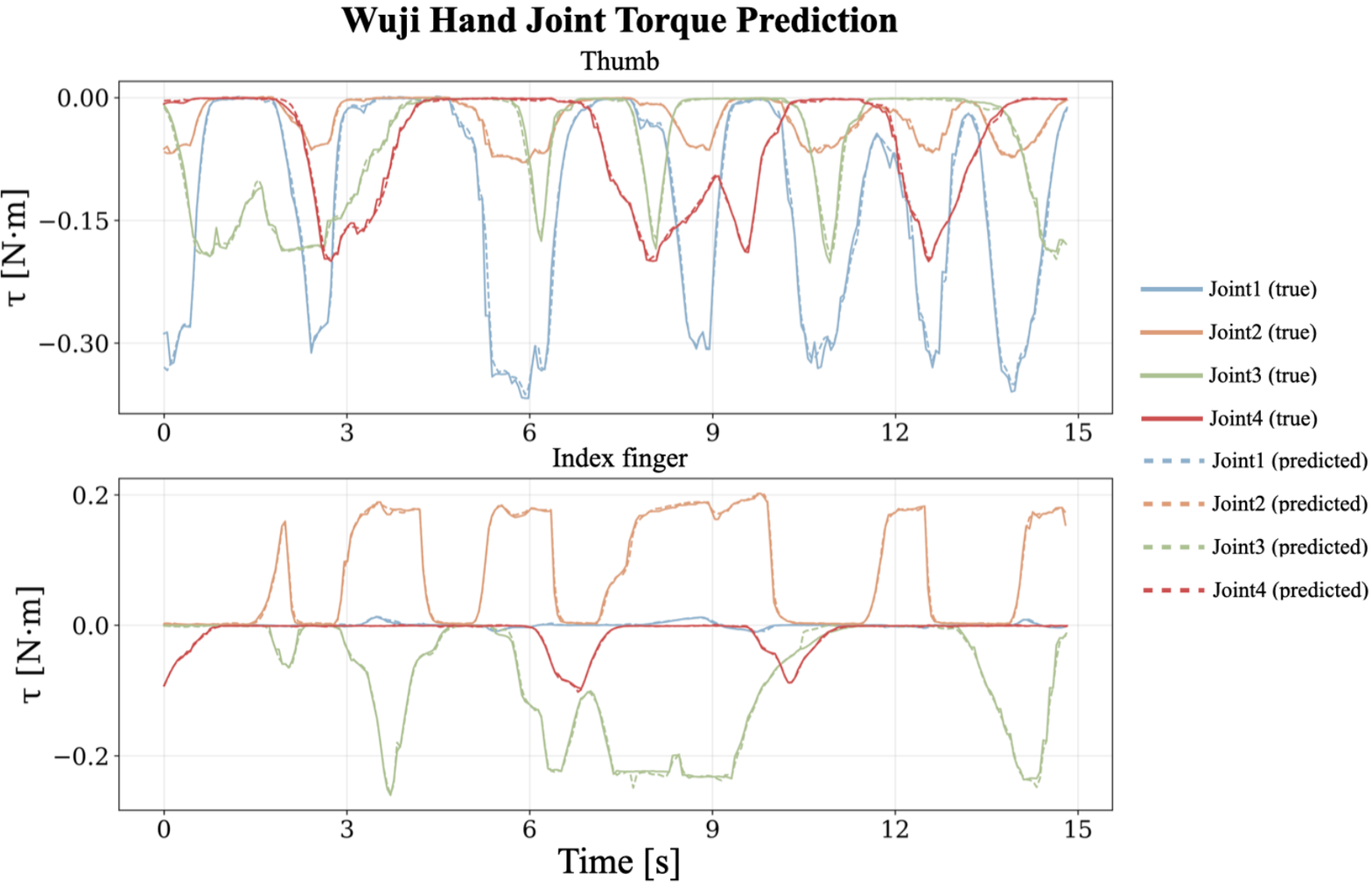}
    \includegraphics[width=0.48\linewidth, trim=0 50 50 250]{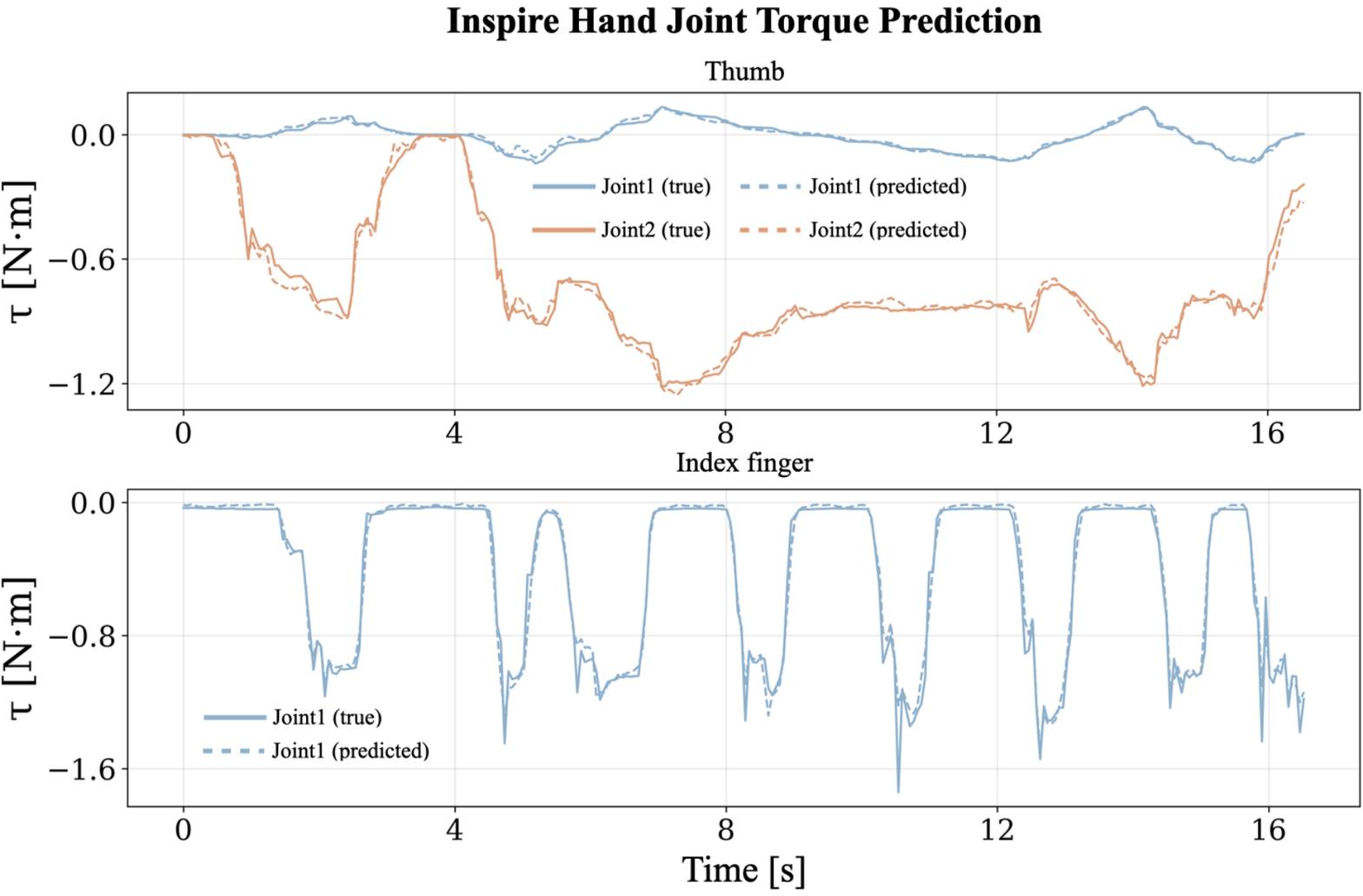}
    \caption{Qualitative torque estimation on the Inspire (left) and Wuji (right) hands. Predicted torques from $G_h$ are overlaid against ground-truth labels across a representative trial, showing tracking fidelity under natural finger motion.}
    \label{fig:qual_torque_pred}
\end{figure}

\paragraph{Ablation summary.}
We ablate history length $n$, model type, and input features on both hands (80/10/10 split); key findings are reported in Tables~\ref{tab:wuji_ablation} and~\ref{tab:inspire_ablation}. Three results drive the final design choices. (i)~For Wuji, an LSTM achieves $2.8\times$ lower MSE than a size-matched MLP ($0.021$ vs.\ $0.060$) because Wuji's non-backdrivable actuation imprints temporal dynamics that a flat model cannot capture; $n{=}25$ balances accuracy and latency. MSE continues to improve beyond $n{=}7$ but extending further risks estimation delays incompatible with the $20\,\mathrm{Hz}$ control rate. (ii)~For Inspire, MLP, LSTM, and CNN reach identical MSE at every size tested, so the smallest MLP is used; $n{=}7$ is adopted (optimum at $n{=}4$, but $n{=}7$ shows no degradation and better suppresses low-frequency noise). (iii)~On both hands, the effort proxy $\tilde{\boldsymbol{\tau}}^{(h)}$ is the dominant feature where adding velocity without effort \emph{increases} error; so $[q, \dot{q}, \tilde{\boldsymbol{\tau}}^{(h)}]$ is used for both hands Final configurations are given in Table~\ref{tab:final_arch} and qualitative analysis of these final configuration are shown in Fig.~\ref{fig:qual_torque_pred}.


\begin{table}[h]
\centering
\caption{Wuji ablations (non-thumb fingers pooled).
  \emph{Left}: history length (MLP-large, $[q,\dot{q},\tilde{\boldsymbol{\tau}}^{(h)}]$).
  \emph{Right}: model type ($n{=}7$, small size). Normalized test MSE.}
\label{tab:wuji_ablation}
\begin{tabular}{cc@{\hspace{2em}}llc}
\toprule
\multicolumn{2}{c}{History length} & \multicolumn{3}{c}{Model type} \\
\cmidrule(r){1-2}\cmidrule(l){3-5}
$n$ & MSE$\downarrow$ & Type & Params & MSE$\downarrow$ \\
\midrule
 1  & 0.117 & MLP  &  7{,}652 & 0.060 \\
 5  & 0.038 & CNN  & 16{,}516 & 0.034 \\
 7  & 0.031 & LSTM & 53{,}508 & \textbf{0.021} \\
15  & 0.024 &      &          &       \\
\textbf{25} & \textbf{0.018} & & & \\
40  & 0.017 &      &          &       \\
\bottomrule
\end{tabular}
\end{table}


\begin{table}[h]
\centering
\caption{Inspire ablations (non-thumb fingers pooled).
  \emph{Left}: history length (MLP-large, $[q,\dot{q},\tilde{\boldsymbol{\tau}}^{(h)}]$).
  \emph{Right}: model type ($n{=}7$, small size). Normalized test MSE.}
\label{tab:inspire_ablation}
\begin{tabular}{cc@{\hspace{2em}}llc}
\toprule
\multicolumn{2}{c}{History length} & \multicolumn{3}{c}{Model type} \\
\cmidrule(r){1-2}\cmidrule(l){3-5}
$n$ & MSE$\downarrow$ & Type & Params & MSE$\downarrow$ \\
\midrule
 1  & 0.051 & MLP  & ${\approx}2{,}200$ & 0.022 \\
 3  & 0.027 & CNN  & ${\approx}2{,}200$ & 0.022 \\
 4  & 0.021 & LSTM & ${\approx}2{,}200$ & 0.022 \\
\textbf{7}  & \textbf{0.021} &      &                    &       \\
10  & 0.023 &      &                    &       \\
\bottomrule
\end{tabular}
\end{table}


\begin{table}[h]
\centering
\caption{Final torque-regressor configurations.
  All groups pool data across fingers sharing the same actuator type. $n$: history length; $d_{\mathrm{in}}$: features per step (LSTM) or total flat input (MLP); $d_{\mathrm{out}}$: predicted torques.}
\label{tab:final_arch}
\begin{tabular}{lllccc}
\toprule
Hand & Group & Architecture & $n$ & $d_{\mathrm{in}}$ & $d_{\mathrm{out}}$ \\
\midrule
Wuji    & j3/j4, all fingers  & LSTM $h{=}64$, $L{=}2$ & 25 & 3\,(seq) & 1 \\
Wuji    & j1+j2               & LSTM $h{=}64$, $L{=}2$ & 25 & 6\,(seq) & 2 \\
Inspire & non-thumb (4 fingers) & MLP $[64{\to}32{\to}1]$      &  7 & 21\,(flat) & 1 \\
Inspire & thumb               & MLP $[256{\to}128{\to}64{\to}2]$ &  7 & 42\,(flat) & 2 \\
\bottomrule
\end{tabular}
\end{table}

\section{MARC vs SDiT}
Both policies share the hyperparameters listed in Tab.~\ref{tab:hyperparams}.
The sole architectural difference is MARC's structured visual masking Tab.~\ref{tab:marc_sdit_config}. Ablation study on MARC as shown in Tab.~\ref{tab:marc_mask_ablation}. Removing fingertip force results in the largest performance drop (83.3\% to 39.0\%), confirming that calibrated contact feedback is the most critical input modality, whereas removing fingertip position has a milder effect (66.7\%). Reducing the action horizon from 150 to 48 steps also significantly degrades performance (44.3\%), suggesting that longer action chunks are better suited for contact-rich manipulation. Visual masking probability requires careful tuning: moderate masking (patch 0.15 \& camera 0.1) preserves reasonable performance (66.7\%), whereas aggressive masking (patch 0.3 \& camera 0.1) over-suppresses visual cues and collapses to 33.3\%.

\begin{table}[t]
\centering
\caption{Configuration differences between sDiT and MARC. All other hyperparameters are identical: batch size 64, 60k training steps, $H{=}150$ action horizon, 149 executed steps, 16 inference steps, $240{\times}320$ resize / $216{\times}288$ crop, ImageNet-1K backbone, Beta noise sampling, lr $1.4{\times}10^{-4}$.}
\label{tab:marc_sdit_config}
\begin{tabular}{lcc}
\toprule
Parameter & sDiT & MARC \\
\midrule
Image latent masking      & No   & Yes  \\
Patch dropout prob.\      & 0.00 & 0.1 \\
Full-camera dropout prob.\ & 0.00 & 0.05 \\
\bottomrule
\end{tabular}
\end{table}

\begin{table}[h]
\centering
\caption{Shared training and architecture hyperparameters for both policies.}

\begin{tabular}{ll}
\toprule
\textbf{Hyperparameter} & \textbf{Value} \\
\midrule
Observations & RGB, depth, template \\
Observation steps & 2 \\
Action horizon $H$ & 150 \\
Executed steps per chunk & 149 \\
Inference steps & 16 \\
Training steps & 60{,}000 \\
Batch size & 64 \\
Learning rate & $1.4 \times 10^{-4}$ \\
Noise sampling & Beta \\
Backbone weights & ImageNet-1K \\
Backbone normalization & Group normalization \\
RGB encoders & Separate per camera \\
Input resolution & $240\times320 \to 216\times288$ (resize-crop) \\
Proprioceptive state dim & 69 \\
\bottomrule
\label{tab:hyperparams}
\end{tabular}
\end{table}

\begin{table}[t]
\centering
\caption{Ablation study of MARC. We vary the action horizon, input modalities (fingertip position and force), and per-patch and per-camera visual masking probabilities. Success rate (\%) is averaged over six rollouts per hand.}
\label{tab:marc_mask_ablation}
\small
\renewcommand{\arraystretch}{1.15}
\setlength{\tabcolsep}{8pt}
\begin{tabular}{l ccc c}
\toprule
\textbf{Variant} & Inspire & Wuji$_{12}$ & Wuji & \textbf{Avg}\\
\midrule
\textbf{MARC} & \textbf{1.0} & \textbf{0.83} & \textbf{0.83} & \textbf{0.89}\\
\midrule
\multicolumn{5}{l}{\emph{Prediction horizon}}\\
\quad horizon 48 & 0.33 & 0.33 & 0.67 & 0.44\\
\midrule
\multicolumn{5}{l}{\emph{Input modality}}\\
\quad w/o fingertip position        & 0.5 & \textbf{0.83} & 0.67 & 0.67\\
\quad w/o fingertip force and spatial torque descriptor          & 0.67 & 0.33 & 0.17 & 0.39\\
\midrule
\multicolumn{5}{l}{\emph{Visual augmentation (patch / camera noise)}}\\
\quad patch 0.15,\ camera 0.1  & \textbf{1.0} & 0.5 & 0.5 & 0.67\\
\quad patch 0.3,\ camera 0.2   & \textbf{1.0} & 0.5 & 0.17 & 0.56\\
\quad patch 0.3,\ camera 0.1   & 0.5 & 0.17 & 0.33 & 0.33\\
\bottomrule
\end{tabular}
\end{table}

\section{Controller Parameters and Stability}
\label{app:controllers}

\paragraph{Steady-grasp regulator.}
The grip target $\tau^\star + k_a\|\mathbf{a}\|$ in Eq.~\eqref{eq:transport-grasp} is capped at $\tau_\mathrm{max}$ to prevent crushing under acceleration spikes: $\tau_\mathrm{tgt} = \min(\tau^\star + k_a\|\mathbf{a}\|,\,\tau_\mathrm{max})$. The projection $\mathrm{clip}_{[0,\Delta q^+]}$ enforces two hard constraints: the lower bound prevents the command from retreating past $\mathbf{q}_g$, and $\Delta q^+$ is a per-hand crush margin set at calibration. Outside the projection boundary, the torque error $\tau_\mathrm{tgt} - |\boldsymbol{\tau}^i|$ contracts geometrically at rate $(1-\kappa)$ per step, so the unique fixed point $|\boldsymbol{\tau}| = \tau_\mathrm{tgt}$ is asymptotically stable; inside the boundary, the clip provides a hard safety guarantee independent of gain.

\paragraph{Handover progress ODE.}
Let $V(\beta)=(1-\beta)^2$. Then 
\begin{equation}
    \dot{V} = -2K\bigl[\|\mathbf{F}_\Delta\|^2 - F_d^2\bigr]_+ V \;\leq\; 0,
\end{equation}
so $[0,1]$ is forward-invariant and $\beta\to 1$ exponentially under any sustained pull with $\|\mathbf{F}_\Delta\|^2 > F_d^2$. Under constant input the solution of Eq.~\eqref{eq:beta-ode} is $\beta(t) = 1 - \exp\!\bigl(-K(\|\mathbf{F}_\Delta\|^2-F_d^2)\,t\bigr)$, giving time-constant $\mathcal{T}=[K(\|\mathbf{F}_\Delta\|^2-F_d^2)]^{-1}$ set by pull magnitude. A startup grace period $T_\mathrm{grace}$ forces $\dot\beta=0$ to absorb the phase-transition wrench transient. Termination requires $\beta>\beta^\star$ sustained for debounce window $T_d$.

\begin{table}[h]
\centering
\caption{Steady grasp parameters used for stacked cups, cookie, and egg. Handover parameters used for all parameters}
\label{tab:controller-params}
\small
\begin{tabular}{llll}
\toprule
Symbol & Description & Value & Unit \\
\midrule
\multicolumn{4}{l}{\textit{Steady-grasp}} \\
$\tau^\star$         & Resting grip target     & 0.10 & N$\cdot$m \\
$\tau_\mathrm{max}$  & Grip ceiling            & 0.15 & N$\cdot$m \\
$k_a$                & Acceleration boost gain & 0.05 & N$\cdot$m\,s$^2$/m \\
$\kappa$             & Integrator gain         & 0.05 & rad/(N$\cdot$m) \\
$\Delta q^+$         & Crush margin            & 0.05 & rad \\
\midrule
\multicolumn{4}{l}{\textit{Handover}} \\
$K$                  & Progress gain           & 0.02 & $({\rm N}^2\,{\rm s})^{-1}$ \\
$F_d$                & Force deadband          & 8.0  & N \\
$\beta^\star$        & Termination threshold   & 0.98 & -- \\
$T_d$                & Debounce window         & 0.25 & s \\
$T_\mathrm{grace}$   & Startup grace period    & 0.50 & s \\
$\delta$             & Open-offset fraction    & 0.60 & -- \\
\bottomrule
\end{tabular}
\end{table}

\begin{figure}
    \centering
    \includegraphics[width=0.49\linewidth, trim=100 0 100 0]{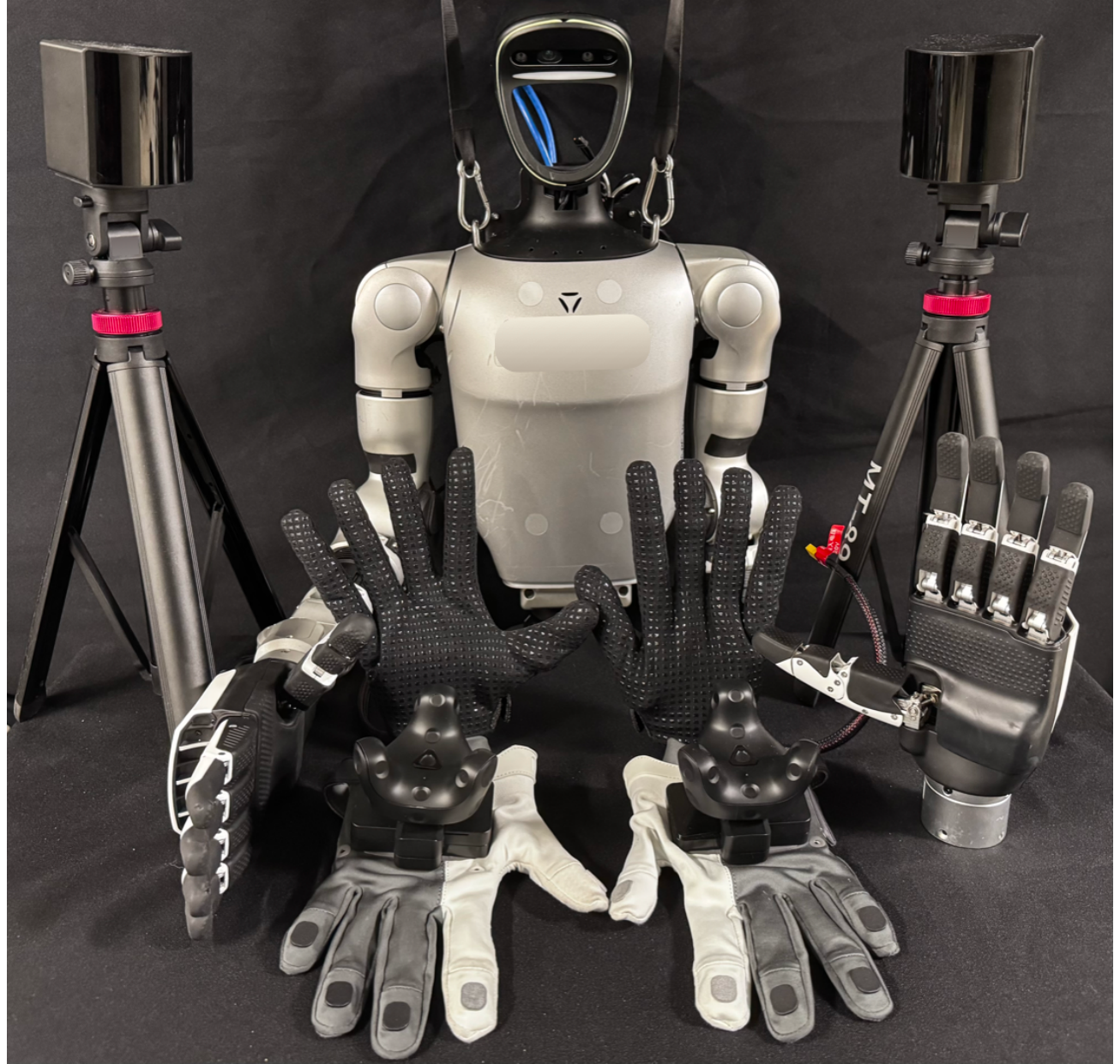}
    \includegraphics[width=0.45\linewidth, trim=0 0 0 0]{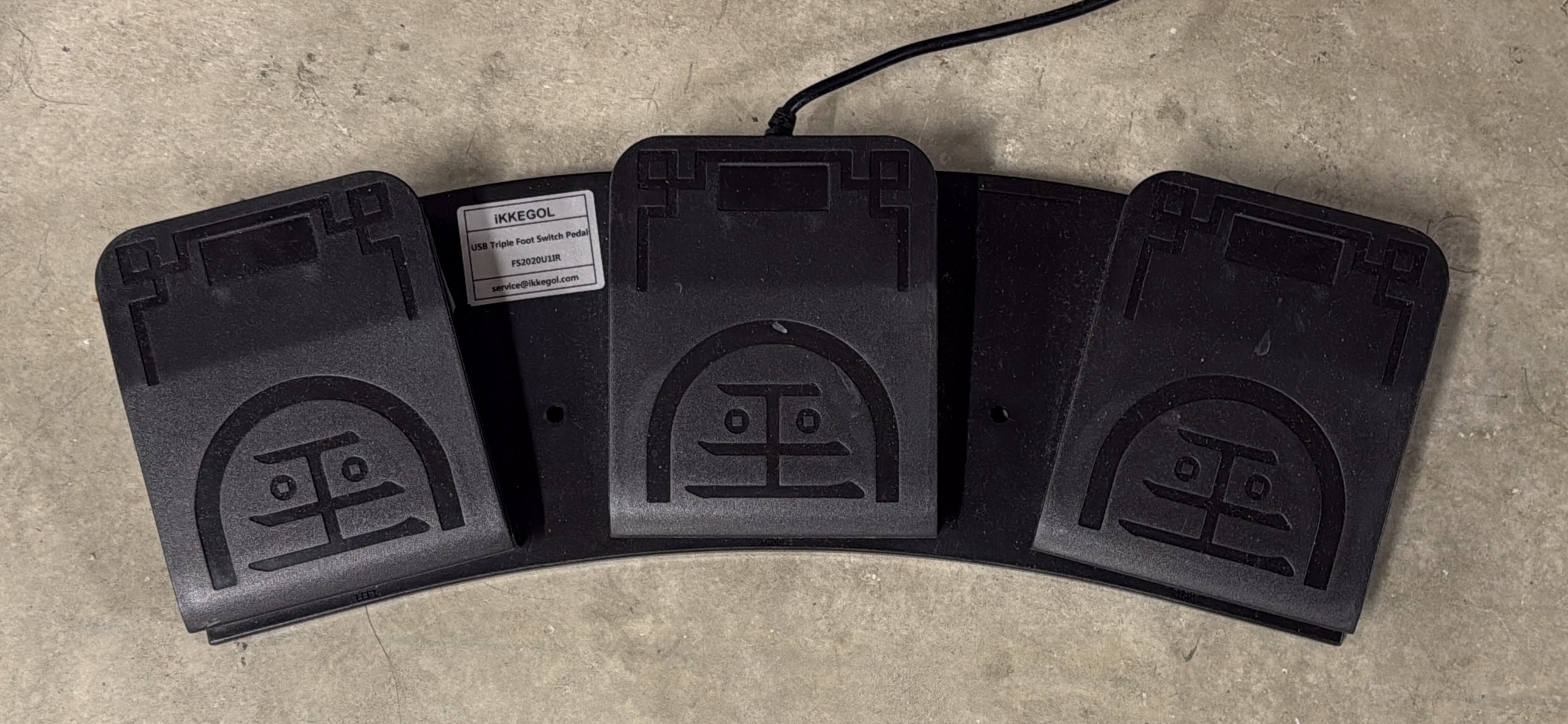}
    \caption{Teleoperation setup using Inspire and Wuji hands, with Vive trackers for end-effector tracking, Manus gloves for dexterous finger motion capture, and a foot pedal as a clutch for operator repositioning.}
    \label{fig:teleop}
\end{figure}

\section{Long horizon tasks}

\subsection{Few examples of VLM Prompt Templates and Outputs}
\label{app:prompts}

\begin{promptcard}{Place star fruit — VLM prompt \& output}

\textbf{\small Prompt}
\begin{lstlisting}[style=inner]
Detect the starfruit and the basket in the image.

Label '0': The bounding box of the starfruit to pick, as [u_min, v_min, u_max, v_max] in pixel coordinates.
Label '1': The bounding box of the basket, as [u_min, v_min, u_max, v_max] in pixel coordinates.

The answer should strictly follow the JSON format:
[
  {"point": [u_min, v_min, u_max, v_max], "label": "0", "type": "bounding_mask"},
  {"point": [u_min, v_min, u_max, v_max], "label": "1", "type": "bounding_mask"}
]

Note: All values are raw numerical floats in pixel coordinates (origin top-left, u = horizontal right, v = vertical down).
Do not use string equations in the arrays.
\end{lstlisting}

\medskip\hrule\medskip

\textbf{\small Output}
\begin{lstlisting}[style=inner]
[
  {"point": [613, 449, 726, 696], "label": "0", "type": "bounding_mask"},
  {"point": [0, 361, 314, 995], "label": "1", "type": "bounding_mask"}
]
\end{lstlisting}

\end{promptcard}

\begin{promptcard}{Handover Bun — VLM prompt \& output}

\textbf{\small Prompt}
\begin{lstlisting}[style=inner]
Detect the top bun and the plate in the image.

Label '0': The bounding box of the top bun to pick, as [u_min, v_min, u_max, v_max] in pixel coordinates.
Label '1': The bounding box of the plate to place the top bun on, as [u_min, v_min, u_max, v_max] in pixel coordinates.

The answer should strictly follow the JSON format:
[
  {"point": [u_min, v_min, u_max, v_max], "label": "0", "type": "bounding_mask"},
  {"point": [u_min, v_min, u_max, v_max], "label": "1", "type": "bounding_mask"}
]

Note: All values are raw numerical floats in pixel coordinates (origin top-left, u = horizontal right, v = vertical down).
Do not use string equations in the arrays.
\end{lstlisting}

\medskip\hrule\medskip

\textbf{\small Output}
\begin{lstlisting}[style=inner]
[
  {"point": [477, 345, 630, 567], "label": "0", "type": "bounding_mask"},
  {"point": [657, 32, 932, 300], "label": "1", "type": "bounding_mask"}
]
\end{lstlisting}

\end{promptcard}

\begin{promptcard}{3-cup pyramid — VLM prompt \& output}

\textbf{\small Prompt}
\begin{lstlisting}[style=inner]
First, detect the initial pile of cups and provide its absolute 3D starting coordinate. Then, provide the exact 3D displacements needed for the next two placements to construct a 3-cup pyramid.

Label '0': absolute starting 2D coordinate [u, v] of the initial pile.
Label '1': relative displacement [dx, dy, dz], next cup to the left.
Label '2': relative displacement [dx, dy, dz], final cup on top.

Format: raw numerical floats.
Convention: right-hand rule, x forward, y left, z up.
\end{lstlisting}

\medskip\hrule\medskip

\textbf{\small Output}
\begin{lstlisting}[style=inner]
[
  {"point": [504, 503],         "label": "0", "type": "absolute"},
  {"point": [0.0, 0.12, 0.0],  "label": "1", "type": "relative_displacement"},
  {"point": [0.0, 0.06, 0.15], "label": "2", "type": "relative_displacement"}
]
\end{lstlisting}

\end{promptcard}

\end{document}